\documentclass[12pt,leqno]{amsart}
\usepackage[T1]{fontenc}
\usepackage[utf8]{inputenc}
\usepackage{orcidlink}
\usepackage[margin=1in]{geometry}
\usepackage{setspace}
\usepackage{parskip}
\usepackage[protrusion=false,expansion=false]{microtype}
\usepackage{hyperref}
\usepackage{url}
\usepackage{natbib}
\usepackage{booktabs}
\usepackage{enumitem}
\usepackage{abstract}

\hypersetup{
  colorlinks=true,
  linkcolor=black,
  citecolor=black,
  urlcolor=black,
  pdftitle={
  Rethinking Publication: Certifying Knowledge and Contribution in AI-Enabled Research
  },
  pdfauthor={}
}

\bibpunct{(}{)}{;}{a}{,}{,}

\begin{document}


\def\proof{\noindent {\it Proof. $\, $}}
\def\endproof{\hfill $\Box$ \vskip 5 pt }
\title{Rethinking Publication: A Certification Framework for AI-Enabled Research}
\author[Lu]{Yang Lu \orcidlink{0009-0006-0981-7211}}
\author[Karanjai]{Rabimba Karanjai \orcidlink{0000-0002-6705-6506}}
\author[Xu]{Lei Xu \orcidlink{0000-0002-7662-2119}}
\author[Shi]{Weidong Shi \orcidlink{0000-0002-1994-4218}}
\date{\today}
\address{Department of Computer Science, University of Houston, Houston, Texas}
\email{ylu17@central.uh.edu}
\maketitle

\begin{abstract}
\noindent\textbf{Background:} AI research pipelines now generate a
substantial and growing fraction of publishable academic output,
including work that meets existing peer review standards for quality
and novelty. The publication system was designed when human authorship
of all research was universal and has no principled basis for evaluating
knowledge whose producer is automated.

\noindent\textbf{Objective:} This paper proposes a two-layer
certification framework that separates knowledge quality assessment from
human contribution grading, enabling the publication system to handle
pipeline-generated work consistently without requiring new
institutions.

\noindent\textbf{Methods:} Normative-analytical methodology,
conceptual analysis, framework design against four explicit constraints,
and dry-run validation against two representative submission cases
spanning the key attribution scenarios.

\noindent\textbf{Findings:} The proposed framework grades contributions
as Category~A (pipeline-reachable), Category~B (human direction required
at identifiable stages), or Category~C (beyond current pipeline reach at
the formulation stage). Dedicated benchmark slots for fully-disclosed
automated research serve simultaneously as a transparent publication
track and a living calibration instrument for reviewer judgment. The
contemporaneous standard\footnote{\textbf{Contemporaneous standard}: contribution grading assessed against what the best available pipeline could generate at the time of submission, assessed against what the submitter knew at time of submission, not retroactively adjusted as capability advances.} applies contribution grading relative to
pipeline capability at the time of submission. Dry-run validation
confirms the framework certifies knowledge correctly while tolerating
irreducible attribution uncertainty.

\noindent\textbf{Conclusions:} Publication has always made two
certifications simultaneously: that the knowledge holds, and that
a human made it. The pipeline has, for the first time,
separated them. Decoupling certification from attribution
recognizes a structural change that has already occurred. The framework is implementable using existing editorial
infrastructure, functions despite irreducible attribution
uncertainty, and grounds the recognition of human frontier
contribution in its epistemic property rather than in human origin
alone.
\end{abstract}

\vspace{12pt}
\hrule
\vspace{12pt}

\section{Introduction}

\subsection{The Current Situation}

AI research pipelines now generate output that meets peer-review standards at scale across multiple disciplines as an accelerating operational reality \citep{lu2024, yamada2025}. In fully automated mode, a pipeline identifies literature gaps, generates and ranks hypotheses, designs and executes analysis, and produces manuscript-formatted output indistinguishable in format and methodological compliance from human-authored work \citep{analemmaFARs}. This mode is already operational in systematic literature review and meta-analysis \citep{sami2025systematicliteraturereviewusing}, materials science \citep{Szymanski2023}, generative drug discovery \citep{Zhavoronkov2026}, and computational biology and genomics, where agentic frameworks are transforming autonomous pipelines across tasks from sequence analysis to hypothesis generation \citep{Gao2024EmpoweringBD}. To make this concrete: a pipeline systematic review of dietary interventions for Type 2 diabetes reads several hundred papers, extracting intervention types, sample sizes, outcome measures, and effect sizes. It synthesizes these into a meta-analytic summary with heterogeneity statistics and produces a manuscript formatted to the target journal's conventions: correctly cited, methodologically sound, delivered within hours. A pipeline, given a corpus of digitized historical newspapers, identifies all mentions of a named entity, traces frequency and co-occurrence patterns over time, and produces a structured summary: a task that previously demanded months of manual research effort \citep{Ehrmann2023}.

The empirical scale of this shift is documented, though its full extent
is structurally obscured by the same problem the paper addresses.
\citet{Liang2024} found that a substantial and growing proportion of AI
conference peer reviews showed evidence of AI modification,
unreported and undetected by the venues that received them.
Detection-tool accuracy itself is limited: OpenAI's own classifier
correctly identifies only 26\% of AI-written texts while flagging 9\%
of human-written texts \citep{StokelWalker2023}. \citet{Guo2023}
identified statistically significant increases in AI-associated
linguistic patterns in biomedical preprints following the release of
large language models. These measurements share a common limitation:
they capture what the detection apparatus can see.
The fraction of pipeline-generated papers that pass peer review
undetected is, by definition, unobservable under the current system,
which is precisely why the certification framework proposed here does
not depend on detection.

\subsection{The Specific Disruption}

Publication in the academic knowledge enterprise is not primarily a
dissemination mechanism. It is a certification mechanism. When a paper
is published, the publication certifies two things simultaneously: that
the knowledge claim meets the community's quality standards, and that a
qualified human made a contribution sufficient to claim the work. These
two certifications have always been performed in the same act, by the
same reviewers, using the same process, because they could be: when
humans wrote everything, certifying the knowledge and certifying the
contribution were indistinguishable.
\footnote{Publication performs several functions alongside knowledge
certification: credentialing, community membership, resource
allocation. The framework addresses only the certification function,
which is the one the pipeline has separated from the others. The social
and institutional functions remain important and are not this paper's
subject.}

A scope clarification before the argument proceeds. Whether pipeline
output constitutes knowledge in a deep philosophical sense is
contested; this paper proceeds from the narrower observation
that peer review currently certifies much pipeline output as meeting
the community's quality threshold. The framework addresses what the
publication system does, not what knowledge fundamentally is.

Automated pipelines separate these for the first time. A pipeline can
produce output that peer review certifies as valid, novel,
methodologically sound, and worth the community's attention, and the
publication system has no principled basis for treating this as a
reviewer error. If the knowledge is valid, the reviewers did not err.
They correctly identified work that meets quality standards. The
problem is not reviewer failure. It is system design: the system was
not designed to certify knowledge independently of person, and it has
no mechanism for doing so. The deeper implication is not merely that
the system needs updating. It is that the pipeline reveals what
publication has always been at its foundation: not only a statement
about persons, but also a statement about knowledge. The second
function, this paper argues, can be performed independently
of the first, and the structural change suggests it should be. The pipeline has made this true in practice; the question this paper addresses is how policy might acknowledge that change.

A methodological note before proceeding. The characterization of
pipelines as autonomous producers throughout this paper is an
analytical simplification, not an empirical claim about typical use.
In practice, researchers direct pipelines continuously, and the
framework's Category~B grade (\S3.2) is designed to recognize those
human-directed cases. Fully automated mode is treated first because
it is the worst case for any certification system: a framework that
handles it would handle the intermediate cases. 
The concern is not that researchers will be replaced, but that the current system has no principled way to evaluate pipeline-assisted work.

\subsection{Why Existing Responses Fail}

Detection-based tools address the wrong layer \citep{WeberWulff2023}.
A text that reads like human writing may be pipeline-generated; a text
that reads like AI output may be substantially directed and revised by a
human researcher. Writing style is not the contribution question. The
knowledge contribution is the question. A pipeline-generated paper that
reads as human-written and makes a genuine frontier contribution is
better for the knowledge enterprise than a human-written paper that
makes a low-bar contribution and reads awkwardly. A system that detects the former and accepts the latter is misaligned with what it is meant to certify.

Disclosure-only policies have empirically failed at scale. Research
analyzing over 75,000 academic papers published since 2023 found that
approximately 0.1\% explicitly disclosed AI use, despite 70\% of
journals having adopted mandatory disclosure policies
\citep{HeBu2025}. The \citet{Liang2024} finding, unreported AI
modification across four major AI conferences, confirms the same
pattern at the venue level. The structural reason is clear: under current policy,
disclosure has costs, stigma, uncertain liability, and no
benefits. Nothing in the current system rewards transparent disclosure
or protects the researcher who discloses honestly from the same
competitive pressure as the researcher who does not.

Prevention has limited viability as a policy position. Researchers have already adopted
pipelines because they work, faster, better literature coverage, more
methodologically consistent outputs. The pace of pipeline adoption among researchers has effectively settled the question of whether before institutional policy has had time to address it. The relevant
question is no longer whether but how.

Preprint repositories have been valuable for priority registration
but are structurally misaligned with the current bottleneck. They were
designed for a generation-constrained world: ideas and writing were
slow, and the repository's function was to accelerate communication of
complete human-authored work. In the AI-native era the bottleneck has
shifted: generation is fast and cheap; selection, verification, and
attribution are the scarce activities. A high-agency researcher using
current pipelines can generate hundreds of candidate research outputs
per year. The implicit authorship model of preprint repositories,
the submitter authored the work, does not accommodate the
idea-originator role that is now the primary location of irreplaceable
human contribution.

This paper does not argue that pipeline output is equivalent to human
intellectual contribution. It argues that the publication system should
certify knowledge quality regardless of origin while separately and
explicitly recognizing the specific human cognitive act that pipelines
cannot replicate, and that these two functions, currently conflated,
can be productively separated.

\subsection{Contribution of this Paper}

This paper proposes a two-layer certification framework for the
publication system. Layer~1 is quality certification: pipeline-agnostic,
unchanged from current practice. Layer~2 is contribution grading on a
single axis of pipeline-generatability at the time of submission,
producing three grades (A,~B,~C). The framework introduces dedicated
benchmark slots for fully-disclosed automated research, a
contemporaneous standard for contribution grading, and a challenge
mechanism for high-stakes attribution disputes. It is designed to
function without relying on author honesty and to be implementable using
existing editorial infrastructure without new institutions.

Section~2 reviews related work and states the methodology. Section~3
presents the framework. Section~4 validates it against two
representative submission cases. Section~5 discusses reviewer
qualification, equity and institutional implications, scope and
limits, and partial adoption paths. Section~6 concludes.

\section{Background}

\subsection{Existing Responses and Their Limits}

The discourse on AI and academic research has generated different categories
of response. Each addresses an important aspect of the problem; this
paper argues that a structural framing is also needed alongside them.

The question ``is AI-generated work acceptable?'' presupposes that
acceptability is a property of origin rather than content. The
framework proposed here treats what the work contributes as the more
useful question, with origin treated as a separate disclosure matter.

The question ``how do we detect AI-generated writing?'' similarly
addresses the wrong layer. Authorship detection tools operate on
stylistic and statistical properties of text. A pipeline can produce
text with human stylistic properties; a human can produce text with AI
stylistic properties after heavy direction and revision. What matters is
the epistemic property of the contribution, whether it required the
specific cognitive act pipelines cannot yet perform, not the surface
features of the writing.

Individual venue AI-use policies represent a fragmented response.
Stances diverge substantively across major journals, ranging from
disclosure requirements to outright bans on AI-generated text
\citep{StokelWalker2023}, producing inconsistency: different venues
adopt different standards, authors route submissions strategically,
and the policy landscape creates compliance incentives that are
orthogonal to knowledge quality.
No systematic framework exists. The \citet{Liang2024} finding,
substantial AI modification of peer reviews, unreported, is the
empirical demonstration that venue-level disclosure policies without
structural enforcement produce documented non-compliance at scale.


Existing attribution frameworks address the disaggregation of credit but
not the certification of contribution level. The Contributor Roles
Taxonomy (CRediT; \citealt{Holcombe2019}) specifies 14 author roles
including conceptualization, methodology, and investigation, and has
been adopted by major publishers including Elsevier and Springer Nature.
CRediT correctly distinguishes who contributed what, but it provides no
mechanism for assessing whether any given contribution exceeded pipeline
capability. A researcher credited with ``conceptualization'' under CRediT
may have performed a consensus-contradicting abductive move or may have
provided a conventional problem statement the pipeline would have
generated independently. CRediT resolves the credit disaggregation
problem; it does not resolve the certification problem this paper
addresses.

Professional society guidelines similarly address conduct but not
contribution level. The International Committee of Medical Journal
Editors \citep{ICMJE2023} updated its authorship guidelines to state
that AI tools cannot be authors because they cannot take accountability
for the work. This is correct on accountability, but creates an
asymmetry: the human who ran the pipeline takes accountability for
output they may not have directed or verified. The vetting spectrum\footnote{\textbf{Vetting spectrum}: the four contribution levels at which a researcher may engage with pipeline output: idea-originator (provides formulation only), endorser (reviews and accepts responsibility), co-director (iterates with the pipeline at identifiable stages), and full-author (verifies all content).} in
Section~3.1 addresses this gap. The Committee on Publication Ethics
similarly updated its guidance on authorship and AI tools
\citep{COPE2023}. \citet{Hosseini2023} propose that AI contributions be
disclosed in methods sections; \citet{Huang2023} propose a transparency
framework for generative AI in research. Neither addresses whether the
disclosed contribution exceeded what the pipeline would have generated
without it.


\subsection{Methodology}

This paper employs normative-analytical methodology: it identifies a structural problem in the current certification system, specifies the design constraints a solution must satisfy, derives framework elements from those constraints, and validates the framework through scenario analysis and dry-run testing against representative cases. The approach sits within the tradition of normative policy analysis and analytic philosophy of science. It is closer to Rawlsian reflective equilibrium \citep{Rawls1971} than to empirical social science, working back and forth between considered judgments about specific cases and coherent general principles.
This places the paper within the broader social epistemology of science \citep{Goldman1999,Kitcher2001,Longino2002}, which directs attention to community-level critical process rather than to individual cognitive acts in the certification of knowledge. This focus becomes empirically testable when producers can be non-human. The framework addresses only the community-certification function peer review performs and does not depend on a particular account of what knowledge fundamentally is.
The framework does not make empirical predictions; it proposes a structure for a certification system and demonstrates that structure's internal coherence.

Four design constraints govern every framework element, stated
explicitly so readers can evaluate each design choice against them.
First, \emph{minimal reliance on author honesty}: the framework
functions on what reviewers can assess from the work itself, not on
what authors disclose. Second, \emph{objectivity}: grading is based on
the work's epistemic properties, not the author's conduct or process.
Third, \emph{alignment with existing practice}: no current review
question is made obsolete; the framework adds a contribution grading
layer to what reviewers already do. Fourth, \emph{implementability}: no
new institutions required. Equity, that the framework not exacerbate
existing inequalities in who gets to produce certified knowledge, is
addressed in Section~5.2 rather than here because the responses it
requires lie beyond what a certification system can mandate.

\section{Framework}

\subsection{The Decoupling Principle and Its Normative Justification}

The primary design move is conceptual before it is operational. Knowledge certification and attribution are logically distinct functions that the publication system currently performs in the same act. Knowledge certification asks: does this knowledge hold, is it novel, and at what contribution level does it belong? Attribution asks: who claims credit and accepts responsibility for this work? For the first time, the pipeline separates these two questions in practice, and the proposed framework formalizes that separation.

Knowledge certification is performed by reviewers acting on the work's
epistemic properties. The answers depend on domain expertise and access
to the dedicated slot benchmark\footnote{\textbf{Dedicated slot benchmark}: a domain-specific reference produced by running the best-available pipeline configurations against agreed problem sets, with top outputs published in dedicated slots of the venue. Provides a concrete, continuously updated record of what current pipelines can produce in the domain. Specified in \S3.4.}, not on author identity or honesty.
It applies identically whether the submitting party is a human
researcher, a team, an organization, or a disclosed automated pipeline.
Authorship remains a credit claim and an accountability assignment; the
contribution certification comes from pipeline-generatability grading\footnote{\textbf{Pipeline-generatability grading}: assessment by domain expert reviewers of whether a submission exceeds what the best available pipeline could have generated at time of submission. Performed against the dedicated slot benchmark. Does not require author honesty to function.},
not from the presence of human names. Authorship has historically
bundled these three functions precisely because they could not be
separated \citep{Biagioli2003}; the pipeline separates them for the
first time.

\textit{The social good argument.} The deepest objection to this
decoupling\footnote{\textbf{Decoupling}: the primary design move of the framework, separating knowledge certification (does the knowledge hold, is it novel, at what contribution level) from attribution (who claims credit and accepts responsibility).} is normative: the knowledge enterprise is a human enterprise,
and its products should reflect human intellectual activity. The
knowledge enterprise is human in the sense that humans created and
maintain it for human benefit. If a pipeline generates knowledge that
advances understanding, refusing to admit it on grounds of origin
treats the question of origin as more important than the question of content. Category~C exists to protect the class
of knowledge pipelines cannot yet generate, the class most at risk
of undercertification as capability advances.

\textit{The consumer perspective.} Knowledge consumers,
practitioners, policymakers, researchers building on prior work,
students, have one primary need from the publication system: quality
assurance. They need to know whether they can trust the knowledge, not
who or what produced it. A cancer treatment that works, works; a system
that certifies quality first, regardless of origin, serves consumers
more directly than one that certifies human contribution first and
struggles to accommodate the pipeline's output.

The argument above establishes that valid knowledge is valid regardless
of its origin. It does not establish that who produces knowledge is
irrelevant. The diversity of knowledge producers, across communities,
languages, institutional positions, and epistemological traditions, is
constitutively valuable: it broadens the range of questions asked and
anomalies noticed. 


\textit{Historical precedent: the Erdős model.} The idea of separating who poses a problem from who solves it is not new. The mathematician Paul Erdős co-authored more than 1,500 papers by sharing problems with collaborators around the world and letting them do the development work \citep{Hoffman1998}. What is genuinely new in the pipeline era is the scale at which this happens, the absence of any expectation that the executor will receive credit, and the fact that ideas can be deposited now and developed years later when capability catches up. The comparison is useful precisely because it breaks down at one specific point. Erdős's collaborators were credited as co-authors. Pipelines make no such claim. That difference dissolves the awkward question of executor credit and leaves the originating idea as the contribution current pipelines cannot replicate.

\textit{The vetting spectrum.} A researcher contributes at different
levels. At the \emph{idea-originator} level, the researcher provides
the formulation without claiming to have verified execution, pipeline
errors are bounded to the execution layer. At the \emph{endorser}
level, the researcher has reviewed the output and accepts
responsibility for its quality. At the \emph{full-author} level, the
researcher has verified all content and takes traditional authorship
responsibility. A fourth level, \emph{co-director}, applies to
iterative pipeline use, now the dominant mode of frontier AI-assisted
research. The co-director provides initial direction, evaluates the
pipeline's response, and iterates, contributing at identifiable
points rather than only at the formulation stage. At every level, what the framework protects is the same: the
specific judgment that constituted the abductive move. The vetting
level determines accountability; the contribution grade determines
certification.

\subsection{Two-Layer Architecture and Contribution Grades}

\textit{Layer~1: Quality certification} is pipeline-agnostic and
unchanged from current practice. The questions are the same questions
reviewers currently ask: Is the methodology sound and correctly applied?
Are the claims supported by the evidence? Is the statistical analysis
correct? Is the writing clear? Does the introduction accurately describe
the current state of knowledge? Is citation coverage adequate? Work
that does not pass Layer~1 is rejected regardless of origin. This layer
explicitly admits pipeline-generated work that meets quality standards.
This is the system's acknowledgment that pipelines can produce
publishable knowledge. One critical addition to several Layer~1
questions: writing quality, comprehensive literature synthesis, and
thorough citation coverage are quality floor criteria, not contribution
indicators. High performance on these dimensions is now consistent with
automated pipeline generation and should not, on its own, be taken as evidence of
frontier human contribution.

\textit{Layer~2: Contribution grading} is pipeline-sensitive and
constitutes the framework's new addition to the review process. Three
existing questions require revision. The \emph{originality} question
should additionally ask whether novelty exceeds what the best available
pipeline could generate from the problem statement and prior literature. The
\emph{impact} question should additionally ask whether the advance
involves redirection of prior consensus, less accessible to automated
generation, or consolidation and confirmation, the pipeline's home
territory. The \emph{introduction quality} question should additionally
ask whether the problem framing reveals a gap that required domain
judgment to identify rather than systematic literature synthesis to
enumerate. Three new questions are required with no current equivalents.
First: could this contribution have been generated by the best available
automated pipeline given the prior literature at the time of submission?
Second: was the research question non-obvious by domain judgment, or
pipeline-reachable by literature synthesis? Third: does the
interpretation of results exceed what automated contextualization would
produce? One additional question completes the layer: does the
manuscript provide sufficient information for the reviewer to perform
this assessment? These questions are asked of the contribution, not
of the author. Whether the author used AI tools is a separate
disclosure question the framework does not settle. For iterative
co-directed pipelines, the contribution question is whether any
distributed human intervention constituted the specific judgment the
framework protects. These questions are written
for reviewers but apply equally as a self-assessment tool: a researcher
who cannot confidently answer ``no'' to the first should reconsider
whether their contribution is Category~B or A rather than C.

The grading system has three levels on a single axis. The axis is
\emph{pipeline reachability}: could the best available pipeline have
generated this contribution, given the prior literature at the time of
submission?

\textbf{Category~A:} The work's problem formulation was within reach of
the best available automated pipeline given the problem statement and
prior literature, regardless of whether the author used AI tools. The work is
valid knowledge, correctly published, but not specially certified as
frontier human contribution.

\textbf{Category~B:} The work required human contribution at identifiable stages that current pipelines handle poorly. The contribution quantum is uncertain but present. This is the framework's most important new recognition: researchers doing genuine human-directed pipeline work,
identifying anomalies and directing execution toward them, currently
have no mechanism to distinguish their contributions from automated
generation in the publication record. Category~B makes that distinction
visible and certifiable.

\textbf{Category~C:} Reviewer consensus that the work's formulation
required a cognitive act automated pipelines cannot yet reliably
perform: the consensus-contradicting abductive move \citep{Peirce1935}.
This means inferring the best explanation for something that does not
fit the existing framework, treating an anomaly as a signal rather
than noise, generating a hypothesis the field's consensus would not have
produced. These are properties of the work, not judgments about the
author's conduct or level of AI use.

A note on what the framework grades. The A/B/C categories are easiest to read as grading the output, that is, what the paper produced. It is more accurate to say they grade the cognitive work the contribution required: specifying a problem and executing within its scope (A), recognizing a confound or anomaly that requires domain judgment (B), or reframing the assumptions the field itself was operating under (C). The framework exists because of AI pipelines, but what it grades is the human cognitive work, not the pipeline. Pipelines serve as the first public benchmark against which such cognitive work can be identified. They are the instrument, not the subject.

\textit{The ceiling condition.} Category~C rests on a contingent
premise: that a class of knowledge exists requiring cognitive acts
current pipelines cannot reliably perform. This premise is not a
symmetric open question, the direction of travel is not neutral.
Pipeline capability is advancing and the Category~C space is narrowing
at different rates across different domains. The ceiling is therefore
better understood not as a single threshold but as a domain-varying,
generation-specific surface: its location differs by field, by problem
type, and by the current capability generation. If pipelines eventually
generate all classes of knowledge including Category~C, the grading
system correctly collapses to a single tier. Until that point,
Category~C operates as a meaningful distinction whose domain-specific
location the calibration database makes tractable to map for the first
time. One implication should be resisted: the directionally shrinking
ceiling does not mean early-career researchers should prematurely
abandon methodological training. Genuine Category~C work requires deep
domain knowledge as a prerequisite, the abductive move is only
possible from within a field, not in advance of it.

One case the framework must handle correctly is the paradigm-founding
contribution: a Category~C insight that opens a research program
subsequently executed largely at Category~A and B. AlphaFold is the
paradigm example \citep{Jumper2021}. The original insight, that
evolutionary co-variation in protein sequences encodes sufficient
structural constraint to determine protein structure to near-experimental
accuracy, was Category~C: it contradicted the field's standing model
and required the abductive move of recognizing that the co-variation
signal was being systematically underused. The follow-on program,
engineering iterations, extension to new molecule types, drug discovery
applications, is correctly Category~A or B. This is the framework
functioning correctly, not devaluing the work. The Category~C
certification belongs to the founding insight; the program's value is
in its scientific fruits, certified at their appropriate level
regardless of their relation to the founding contribution. Researchers
in such follow-on programs may reasonably ask whether their work is
``only'' Category~A. Within the framework's vocabulary, the answer is
largely yes: the founding insight carries the Category~C certification,
and follow-on contributions are graded at A or B as their content
warrants. The framework treats this as accurate description, not
devaluation. Whether this accuracy is welcome is a separate question,
addressed in Section~5.2 on competitive displacement and equity.

A parallel clarification applies to team-based work. Most research has
multiple authors, and the A/B/C judgment applies to the paper's
contribution as a whole, not to any individual author's share. Consider
a three-author paper where one author identified that a widely-used
reference dataset contained a systematic measurement confound, a second
author directed the pipeline's analysis of a corrected dataset, and
the third contributed literature framing. The paper is Category~B:
the confound-recognition is the anomaly-identification act the
framework protects. Authorship credit distributes as in any multi-author
paper and can be recorded in CRediT \citep{Holcombe2019}; the
Category~B certification attaches to the work, not to any one
contributor. This preserves what CRediT does well, disaggregating
credit across roles, while adding what CRediT does not do: certifying
whether the work's contribution exceeded pipeline capability.

\subsection{Contribution Grading in Practice: Three Disciplinary Cases}

A qualification on scope: the consensus-contradicting abductive move
operationalizes frontier contribution within the dominant epistemological
tradition of the current publication system: hypothesis-driven,
anomaly-recognition-based research.
The B/C distinction is a judgment call, not a bright-line test.

\paragraph{\textit{Classification versus value.}} The A/B/C categories describe
the type of cognitive operation a contribution requires, not its value
or publishability.
A Category~A contribution can be enormously valuable: a large-scale
replication study, a definitive normative dataset, a rigorous
cross-population validation.
A Category~C contribution can be wrong.
The framework is a vocabulary for what kind of work something is, not a
ranking of whether it should be published.
Top-tier venues already informally reject routine Category~A work on
significance grounds; the framework makes explicit a distinction venues
already operate on.
The examples below anchor the clearest cases; harder boundary
judgments, papers where reasonable reviewers disagree about A versus
B, are addressed through the calibration database described in
Section~5.

\paragraph{\textit{Scope alignment.}}
Category~A status requires that a paper's claims match what the
pipeline computed.
The failure mode differs by discipline: numeric benchmarks tend to be
scope-aligned because the measurement is the claim, whereas descriptive
outputs require explicit scope narrowing to stay at Category~A, because
linguistic description easily imports interpretive weight the pipeline
did not establish.
A paper that claims more than the pipeline actually showed is not
making a Category~A contribution. Either the human added the larger
claim through a Category~B judgment, or the paper should narrow its
claims to match what the pipeline computed.

\paragraph{\textit{Prompt length.}}
The pipeline invocations sketched below are compressed for illustration.
Production pipeline prompts are typically substantially longer,
specifying datasets, evaluation protocols, tool schemas, and iteration
conditions.
What matters for classification is the cognitive structure of the task
the pipeline is being asked to perform, not the prompt's length.
A ten-thousand-token prompt that mechanically enumerates execution steps
is still Category~A; a fifty-word prompt that names a specific anomaly
and directs investigation toward it can be Category~B.
Prompt elaborateness is not a proxy for contribution depth.

\paragraph{\textit{Closed versus open pipeline runs.}}
A useful way to read the examples below is through the pipeline's mode
of operation.
Category~A contributions are typically produced by \emph{closed}
pipeline runs: the pipeline executes from prompt to manuscript without
mid-course human direction.
Category~B contributions typically arise from \emph{open} pipeline
runs: the pipeline produces output, and the human recognizes something
in that output that the pipeline could not flag as significant,
redirecting the investigation.
Category~B contributions can also arise at the formulation stage,
before any pipeline runs: a researcher who surveys a literature and
recognizes that the field is measuring the wrong things makes the same
cognitive move in a different temporal position.
Category~C examples are historical and sit outside this axis: they
describe the cognitive act that opened a research program, not a
pipeline run.
The abductive move they illustrate is only certifiable in retrospect,
because at the moment it is made most reviewers would classify it as
insufficiently supported; its significance becomes legible only once
the research program it opened has produced results.

\paragraph{\textit{A note to later readers.}} The examples that follow are calibrated to
pipeline capability as of early 2026. A reader in 2036 may find that
tasks here placed in Category~B or C are by then squarely Category~A.
This is the framework operating correctly, not failing: per \S3.5, the
certification attaches to a submission in its contemporaneous context.
A Category~B grading from 2026 records what the contribution took to
produce then, not a standing claim about difficulty.

\subsection*{Computational Biology}

Cross-species transcriptomic concordance between murine models and human
inflammatory disease is condition-specific. Genome-wide correlations
between human leukocyte responses and murine ortholog responses range
from near-zero to $r = 0.08$ across sepsis, burns, and trauma
\citep{Seok2013}, but recover to $r = 0.38$--$0.45$ in specific
conditions when time points are matched \citep{Shay2015}. Pathway-level
rather than genome-wide concordance provides a further methodological
refinement.
Consider a hypothetical researcher who assembles paired human and murine
expression datasets from GEO \citep{Edgar2002} and ArrayExpress
\citep{Athar2019}, matched by inflammatory condition, tissue compartment,
and sampling time point.
An autonomous pipeline executes the retrieval, normalization,
batch-correction, and pathway-level concordance computation, and
produces a systematic catalog reporting, for each comparison, the
concordance magnitude, statistical power, and data coverage, without
claims about which discordances are biologically significant.
Whether the pipeline is implemented by orchestrating standardized
containerized workflows (such as \texttt{nf-core/rnaseq}) or by direct
script generation is an implementation choice that affects reliability,
not classification: both architectures execute the same task from the
problem statement and published corpus.
This is \textbf{Category~A}: the pipeline executes every step given the
datasets and the published corpus, and the paper's claim matches the
catalog's scope.

The \textbf{Category~B} judgment is recognizing that a specific pattern
of pathway-level discordance is mechanistically informative rather than
artifactual.
A pipeline can rank by statistical power.
It cannot determine whether low concordance in a given pathway reflects
genuine cross-species regulatory divergence, a confound such as species
differences in the proportion of neutrophils in whole blood
\citep{Nauseef2023}, or other sample-composition effects that appear
as pathway-level signal but are not regulatory in origin.
Distinguishing these possibilities requires cell-type metadata and
domain knowledge the pipeline does not derive from expression data
alone.
This is not a task that statistical guardrails like kBET or LISI can
solve: those metrics assess whether batch correction succeeded, not
whether a residual cross-species signal reflects genuine regulatory
divergence or a known biological confound.
That judgment is the contribution the framework certifies.

Category~C examples are necessarily drawn from the historical record.
The \textbf{Category~C} example in biology is the Warburg effect.
The standing model held that cancer cells rely on aerobic glycolysis
because their mitochondria are defective \citep{Warburg1956}.
The anomaly was that many tumor cells retain fully functional
mitochondria yet still favor glycolysis in the presence of oxygen.
Warburg's abductive move was to treat this as a feature rather than a
defect: the glycolytic shift supplies biosynthetic precursors that
rapidly proliferating cells need.
This reframing contradicted the standing model and opened the field of
cancer metabolic reprogramming \citep{Hanahan2011}.

\subsection*{Computer Science}
Branch-and-bound search is the engine of exactness in electronic
design automation (EDA), applied across technology mapping, detailed
routing, automatic test pattern generation, floorplanning, gate
sizing, and hardware/software partitioning, each with its own
branching decision and bounding metric.
Recent hybrid LLM/branch-and-bound work clusters around three canonical
patterns (among others) for where the LLM intervenes in the search
loop: LLM-as-branching-heuristic (the LLM proposes the next variable to
branch on), LLM-as-bound-tightener (the LLM proposes valid cuts or
improved bounds), and LLM-as-restart-strategy (the LLM decides when and
how to restart search with new parameters).
Consider a hypothetical study in which a researcher prompts an LLM
agent to survey which (pattern, EDA-sub-problem) cells of this
three-by-six matrix remain unexplored in the literature, and rank the
open cells by expected search-time improvement.
The agent then selects a published baseline with available code and
data in the top-ranked cell, implements the LLM-augmented variant by
transplanting the chosen pattern onto the baseline, benchmarks it
against the original, and produces the manuscript.
This is \textbf{Category~A}.
Survey, matrix-cell enumeration, ranking, and the transplantation of a
published pattern onto a published codebase are executable from the
problem statement and prior literature; no step requires recognizing that a
specific result is mechanistically informative rather than artifactual.

The \textbf{Category~B} judgment arises when the results are in hand.
Suppose the hybrid variant is an LLM-as-bound-tightener applied to
gate sizing, and it reports substantial search-time reduction on a
benchmark of timing-constrained circuits.
A pipeline can report the timing and final objective values.
The \textbf{Category~B} contribution is recognizing that a fraction of the
LLM-proposed bounds silently violate the problem's resource or timing
constraints: the LLM, operating across many bound-generation
invocations with growing context, occasionally drops a hard constraint
(a DSP count, an area budget, a slack requirement) and proposes bounds
that exclude feasible solutions.
The pipeline's reported search-time reduction therefore partially
reflects invalid prunings rather than genuine bound tightening,
because regions excluded by phantom bounds were never examined by the
solver's final feasibility check.
The recognition opens a specific methodological direction: a
verifier-in-the-loop that validates each LLM-proposed bound against an
independent feasibility check before the B\&B solver uses it,
producing a stratified reanalysis that separates genuine LLM
contribution from artifactual gains.
The reformulation is the contribution the framework certifies.

The \textbf{Category~C} example is the Transformer architecture
\citep{Vaswani2017}.
The standing consensus held that modeling sequences required
recurrence: networks that process tokens one at a time and carry state
forward, with long short-term memory \citep{HochreiterSchmidhuber1997}
as the canonical example.
Attention mechanisms existed but were understood as an auxiliary
signal layered on top of this recurrent backbone.
The abductive move was to treat this auxiliary layer as underpinning:
attention alone, with positional encoding providing sequence order,
was sufficient to represent dependencies at any range, and recurrence
could be dispensed with entirely.
No synthesis of the prevailing sequence-modeling literature would have
generated this reframing; it required treating a refinement mechanism
as the primary architecture before large-scale empirical evidence
existed that the resulting system would outperform the recurrent
baselines on the tasks the field cared about.
The research program this opened, the one that produced the large
language models whose capabilities the present paper takes as its
empirical starting point, is a paradigm \textbf{Category~C} contribution of
recent vintage within computer science proper.

\subsection*{Psychology}

Consider a hypothetical study in which a researcher obtains a publicly
available synthetic dataset of adolescent social media usage and depression
markers, generated from assumed population distributions rather than
real behavioral data, and trains a prediction model linking daily
platform use and screen time to depression scores.
The researcher then validates the model against the Adolescent Brain
Cognitive Development (ABCD) Study, a publicly accessible longitudinal
cohort of 11,878 U.S.\ adolescents with real measures of both social
media time and depressive symptoms \citep{Volkow2018, Nagata2025}.
This is \textbf{Category~A}.
The training dataset is synthetic, generated from statistical
assumptions, not real behavior, which is precisely why the ABCD
validation is the scientifically meaningful step: the pipeline answers
a binary empirical question, does the synthetic model generalize to
real longitudinal data, then adjusts model hyperparameters and
produces the paper without further direction.
No anomaly recognition is required; no judgment about which moderator
is theoretically significant is needed.

The same classification applies when the pipeline is an AI agent. Suppose a researcher prompts an agent to produce a catalog of adolescent depression research across developing and Western countries: which populations have been well studied, which are under-represented, and what methods dominate where. This is also \textbf{Category~A}, because the agent can build the catalog from the problem statement alone. Judging which underrepresented populations matter most for theoretical or policy reasons would be the \textbf{Category~B} move.

The \textbf{Category~B} judgment arises when the cross-cultural
comparison is in hand.
Suppose the pipeline reports that the correlation between daily social
media use and depressive symptoms is substantially weaker in
adolescents sampled from some cultural contexts than in the ABCD
cohort, and flags the divergence in the results section.
A pipeline can report the effect-size gap.
A competent pipeline can also run standard measurement-invariance
tests to rule out the possibility that the instrument's items relate
differently to the underlying construct across groups.
The \textbf{Category~B} contribution is recognizing that even when measurement
invariance holds, the effect-size gap may reflect differential
disclosure. Adolescents in some cultural contexts are less likely to
report depressive symptoms on a self-report instrument because of
stigma, family-level pressures, or conventions about disclosing
affective distress \citep{KrendlPescosolido2020}. This is a
response-behavior confound that measurement-invariance testing does
not detect \citep{HenrichHeineNorenzayan2010}.
That recognition redirects the investigation from ``does the effect
generalize'' to ``does the measurement context generalize,'' opening
questions about whether the divergence is substantive or reflects
conditions of data collection that the original pipeline run was not
designed to evaluate.
The reformulation is the contribution the framework certifies.

The \textbf{Category~C} example is cognitive dissonance theory
\citep{Festinger1957}.
The standing model was behaviorism: attitude change follows reward
contingency.
The anomaly, documented by \citet{FestingerCarlsmith1959}, was that
people expressed more favorable attitudes toward a tedious task when
paid less to endorse it.
This was documented but assimilated into the reinforcement framework as
demand characteristics or measurement error.
Festinger's abductive move was to treat the deviation as the signal:
people reduce psychological discomfort by aligning their beliefs with
their behavior, a motivational construct with no place in the
stimulus-response framework.
This opened a research program in attitude change, self-perception, and motivated reasoning.

\subsection{The Dedicated Benchmark Slots}

The dedicated benchmark slots are the framework's most practically
consequential mechanism. They transform pipeline-generatability grading
from an abstract standard into a concrete, continuously updated
reference.

\textit{Design.} Domain-specific problem sets are agreed by field
editorial boards and run against best-available pipeline configurations
in a standardized benchmark competition. Top outputs are published in
dedicated slots with full methodological transparency. This is not open
submission, which creates gaming risk (submitting mediocre pipeline
output to lower the benchmark), and not venue-operated pipelines,
which requires venues to become pipeline operators. The benchmark
competition protocol prevents gaming in both directions and ensures
dedicated slots represent genuine frontier pipeline capability rather
than whatever happens to be submitted.

\textit{Benchmark latency.} The benchmark competition faces a temporal
challenge: field editorial boards operate on annual or biennial cycles,
while frontier AI capability advances on monthly timescales. A benchmark
reflecting capability from twelve months ago will systematically inflate
Category~C certifications during capability transitions, precisely
when the framework is most stressed. The structural response is to
separate governance from operations. Editorial boards retain governance:
they set domain problem sets and quality standards. An independent
infrastructure body handles operations: it runs the benchmark against
the latest available pipeline configurations on a defined schedule,
quarterly is tractable, and publishes updates without waiting for
editorial board deliberation. This allows intellectual standards to be
stable while empirical content tracks the capability frontier.

\textit{Auditability.} The ``best-available pipeline'' standard requires
that benchmark configurations be independently verifiable. A benchmark
run against proprietary black-box systems creates two problems: it
cannot be independently reproduced, and the Category~C ceiling becomes
a function of commercial disclosure choices rather than verified
capability. Participating configurations must therefore be either
open-weight models or proprietary models with published evaluation
specifications sufficient for independent replication.

\textit{Three functions.} First: a transparent publication track for
fully-disclosed automated research, evaluated purely on knowledge
quality, answering the question the current system cannot ask: what
is the best knowledge the pipeline can generate in this domain right
now? Second: a living calibration instrument for reviewer judgment,
making ``could the pipeline have generated this?'' answerable with
reference to a concrete, field-specific record. Third: a public
benchmark making every researcher's competitive position legible, if
a submission does not exceed the dedicated slots, it is Category~A
space; if it exceeds them in identifiable ways, there is a specific
basis for claiming B or C.

\textit{Operational mechanisms.} Calibration database schema,
attribution taxonomy, verifiable records, and the governance
separation between editorial boards and benchmark operations are not
specified in detail here; demonstrating implementability is a
separate, complementary endeavor.

\textit{Governance.} The problem set selection process could be captured
by incumbent researchers who choose domain problems that favor existing
paradigms, systematically underrepresenting the anomalous and
cross-disciplinary formulations where genuine Category~C contributions
are most likely to emerge. The framework's structural response: problem
sets proposed by a rotating editorial committee constituted to include
researchers from outside the dominant paradigms, from subdiscipline
boundaries, emerging subfields, and adjacent disciplines. This is a
modest institutional design requirement, not a new institution.

\textit{Pipeline developer conflict of interest.} Organizations that
simultaneously build frontier AI systems and publish research face a
structural conflict: they help set the Category~A floor through the
benchmark while competing for Category~C certifications at the same
venues. An organization that withholds frontier internal capability from
its benchmark contribution can set the threshold below its own actual
capability. The structural response: organizations whose internal
research pipeline materially exceeds their benchmark contribution must
make an affirmative disclosure of that discrepancy at benchmark
participation.

\textit{Systematic review and SoK papers.} 
These are the paradigm case for dedicated pipeline slots. Comprehensive literature coverage, consistent methodology, and structured synthesis are precisely what automated pipelines do at scale: a pipeline systematic review can match or exceed many human-authored surveys on the metrics the current system uses to accept them. The genre is not eliminated; it is stratified: synthesis execution is Category A; interpretive framing, gap identification, and paradigm-challenging reanalysis belong in the standard track at Category B or C. The framework does not devalue survey work; it distinguishes synthesis execution from interpretive framing.

\subsection{The Contemporaneous Standard}

Pipeline-generatability grading is inherently relative to the technology at the time of submission. A Category~C certification in 2026 reflects that the work exceeded what the best available pipeline could generate in 2026. The same work graded against 2032 pipeline capability might produce a different result. The contemporaneous standard handles this correctly: certification is relative to the technology of the moment. This is not a weakness; it reflects that knowledge contribution is contextual. A researcher who moved the field before the pipeline could have moved it made a genuine frontier contribution, regardless of whether the pipeline eventually catches up. The correct test is always the knowledge state that existed at submission.

Historic papers are not reclassified as pipeline capability improves.
The dedicated slot benchmark of 2026 is archived and permanently
checkable. Retroactive challenge to a Category~C certification asks: was
this work beyond what the 2026 dedicated slot benchmark contained? That
is an answerable question with a documented reference point. The
contemporaneous standard makes retroactive challenge possible and
bounded rather than impossible or unbounded.

\subsection{The Disclosure Regime and Its Limits}

Authors disclose the nature of their pipeline use, what tools were
used, at which stages, and what the nature of the human direction was.
This disclosure is used by reviewers as context for their
pipeline-generatability assessment. A paper disclosing heavy automated
execution at every stage invites more careful Layer~2 scrutiny of
whether the formulation stage involved genuine abductive contribution.
For Category~B contributions specifically, the disclosure should
identify the anomaly that prompted the research direction and the
specific judgment that directed the pipeline. These are the elements a
reviewer must assess to grade the contribution above Category~A; they
cannot be inferred from the manuscript without the researcher naming
them explicitly.

Disclosure is not, however, the sole pillar in the proposed framework.
This is the framework's direct and deliberate response to the
documented failure of disclosure-only policies. The protection against
false attribution is the pipeline-generatability assessment, not the
disclosure: if the work does not exceed the dedicated slot benchmark,
it will not receive Category~C certification regardless of whether the
author claims pipeline-free authorship. If the work does exceed the
benchmark, Category~C is supported by the reviewer's domain judgment
regardless of what the author discloses about their process. The
current system's protection, author honesty, has empirically
failed at scale. The framework realigns the incentive structure:
transparent disclosure is in the researcher's self-interest because it
establishes contribution category, claims priority for the formulation,
and bounds liability to what was actually contributed. For the first
time, transparency serves the researcher's interests rather than working
against them. A practical clarification: disclosure of extensive
pipeline use does not automatically route a submission to dedicated
slots. Dedicated slots are for submissions claiming no human
intellectual contribution. A researcher who used a pipeline heavily but
claims a genuine human contribution, at any vetting level,
submits to the standard track and receives full Layer~2 assessment.

One specific requirement extends the disclosure standard where a
researcher cites a prior public disclosure, a workshop abstract, a
preprint sketch, a conference presentation, as the source of their
research question. The content of that prior formulation must be
provided in the manuscript, not merely a reference to it. A citation
alone does not enable the reviewer to assess whether the formulation was
beyond automated pipeline reach. The specific anomaly identified or
direction proposed must be disclosed for the reviewer to perform their
legitimate Layer~2 assessment.

\subsection{Error Analysis, the Challenge Mechanism, and Self-Correction}

A false positive certifies pipeline-generated work as Category~C when it
should not. The knowledge is valid; the epistemic harm is zero. The
social harm, credit flowing to the wrong actor, is real but
bounded.

A false negative fails to certify genuine frontier work as Category~C.
If the framework systematically under-certifies, the incentive structure
it creates rewards the wrong behavior, which is why the
false-positive preference applies at the B/C boundary.

The challenge mechanism\footnote{\textbf{Challenge mechanism}: retroactive pipeline-generatability test for high-stakes attribution disputes. Run the best available pipeline against the submission-time evidence state and compare against the archived benchmark of that period.} addresses high-stakes attribution disputes, including prize decisions, contested priority, career-progression decisions, and misconduct allegations. A challenger claims that a specific published work was within pipeline reach at the time of submission. The test: run the best available pipeline against the submission-time evidence state using the archived benchmark. If it generates the challenged work, the certification is contested; if not, it is confirmed. The mechanism's existence creates a deterrent: researchers who know Category~C claims can be challenged have an incentive to ensure those claims are genuine. The mechanism extends to one further case. If a challenger presents credible evidence that the submitting organization's internal pipeline capability materially exceeded its benchmark contribution, the certification is subject to the same test. This covers the case where the threshold was effectively set below that organization's actual capability.

Attribution uncertainty is irreducible. A pipeline-generated paper that
genuinely meets Category~C quality criteria would pass the system's
certification, the knowledge has the property Category~C certifies,
and reviewers have correctly identified it. This is tolerable for two
reasons. First, if the work genuinely meets Category~C criteria, the
knowledge is what it is claimed to be regardless of how it was produced;
the attribution error is social, not epistemic. Second, the false
positive rate decreases over time as the benchmark advances and reviewer
judgment calibrates. The framework is not a solution to attribution
uncertainty. It is a system that functions well despite it, more transparently than the current system, with a challenge mechanism for the
cases where uncertainty matters most.

\citet{Goodhart1984}'s Law and self-correction. If the benchmark is
public and researchers know that Category~C requires exceeding it, there
is an incentive to produce work that just clears the visible threshold
without being genuinely frontier, benchmark hugging \citep{Campbell1979}\footnote{\textbf{Benchmark hugging}: gaming the dedicated slot benchmark by producing work that just clears the visible threshold without being genuinely Category~C. Addressed by making certification dependent on reviewer judgment rather than threshold clearance alone.}. This risk is
real and acknowledged. The response has two parts. First, benchmark
hugging still improves on the current floor: clearing a visible pipeline
benchmark required some non-default human engagement. This is a floor,
not a ceiling. Second, Category~C certification depends on reviewer
judgment about the specific character of the contribution, not on
threshold clearance alone. Genuine Category~C work is distinguished by
the specific anomaly identified and the direction pursued, properties
that benchmark clearing does not automatically produce. The benchmark is
one input to the reviewer's judgment, not the sole criterion. Over time,
the system is self-correcting. As benchmark capability advances, the
Category~C bar rises and reviewer calibration adjusts. Community
learning through disputed attributions then refines what genuine
Category~C work looks like in each discipline.

\section{Validation}

\subsection{Purpose and Case Selection}

The validation applies the complete two-layer review architecture to two
representative submission cases chosen to test the framework's handling
of the hardest attribution scenarios. Case~1 is a fully automated
submission with no human attribution claimed, the simplest case,
establishing the baseline. Case~2 is pipeline-generated work in which
the claimed human contribution is a prior public formulation,
the stress-test case, chosen because it directly challenges the
framework's claim to assess epistemic properties without relying on
author honesty or prior-disclosure claims. Both cases are constructed rather
than drawn from real submissions. This is appropriate for
normative-analytical methodology, which validates frameworks by logical
coverage of the case space and internal coherence, not by empirical
prediction for specific real-world instances. Real disclosed pipeline
submissions, as they accumulate in the calibration record, will provide
empirical validation beyond what constructed cases can offer.

\subsection{Case 1: Fully Automated Submission}

\textit{Setup:} an automated pipeline surveys 50 papers, identifies
literature gaps, proposes and ranks research questions, selects the most
impactful, and executes the investigation. Human contribution consists
of providing the papers and the initial prompt. Full disclosure
accompanies the submission.

\textit{Layer~1:} Methodology sound and correctly applied: passes.
Claims supported by evidence consistent with the pipeline's execution:
passes. Statistical analysis methodologically consistent: passes.
Writing quality clear and coherent: passes as quality floor. Introduction
accurately describes the literature: passes as quality floor. Citation
coverage comprehensive: passes. Critical note: writing quality and
citation coverage are quality criteria, not contribution indicators.
High performance on these dimensions is consistent with automated
pipeline generation and should not, on its own, be taken as evidence of frontier
human contribution. Layer~1 verdict: all quality questions pass. The
knowledge is valid.

\textit{Layer~2:} Pipeline-generatability: the submission is the
pipeline output, Category~A confirmed immediately. Problem
formulation: the prompt asked the pipeline to identify and rank gaps;
the formulation is the pipeline's output, not a human's direction.
Interpretation: pipeline-executed, does not exceed automated
contextualization. Revised originality: no novelty beyond pipeline
generation. Revised impact: consolidation and synthesis, the
pipeline's home territory. Layer~2 verdict: Category~A confirmed across
all questions.

\textit{Overall result:} Layer~1 correctly certifies valid knowledge.
Layer~2 correctly grades as Category~A. The work is routed to dedicated
slots. No false positives, no false negatives. The system functions
as designed.

\subsection{Case 2: Pipeline-Generated Work with Prior Public Formulation}

\textit{Setup:} pipeline-generated work disclosing that the research
question originated from a prior public formulation, a problem
statement sketched in a workshop abstract before the pipeline executed
the full investigation. The prior formulation is the stated human contribution. Two
complications test the framework: the prior formulation may itself
be within pipeline reach, and the formulation's novelty is assessed
from its content, not its date of disclosure.

\textit{Layer~1:} Identical to Case~1. Valid knowledge passes the
quality layer regardless of the prior disclosure. The quality of the
knowledge is unchanged by what the formulation's origin was.


\textit{Layer~2:} The reviewer instruction is explicit: assess the work's epistemic properties as read. The prior formulation is context, not criterion. This instruction prevents reviewers from treating the prior disclosure itself as evidence of non-obvious contribution, thereby making the framework gameable by anyone who can produce a prior public sketch. The content of the prior formulation must be provided in the manuscript, not merely cited. The reviewer then assesses whether that formulation reflects a non-obvious anomaly recognition that the pipeline surveying the prior literature would not have produced.

\textit{Sub-case~A:} the prior formulation contains a conventional
research question the pipeline surveying the literature would identify.
Category~A. The prior disclosure establishes temporal priority but not frontier
contribution.

\textit{Sub-case~B:} the prior formulation contains a non-obvious
anomaly recognition, the field assumes X causes Y but the mechanism
inverts in a specific population, and no one has tested this. The
reviewer assesses this formulation as genuinely non-obvious given the literature at the time of disclosure. Content was provided (meeting the disclosure adequacy requirement). Category~B or C, with the false-positive preference applying at the boundary: if uncertain, err toward C.

Three properties of the framework handle the complications this case raises. First, the framework does not rely on prior disclosure to certify human origin. It assesses the formulation's epistemic properties, so whether the prior disclosure is human-authored is not the relevant question. Second, citing a prior disclosure is insufficient: its content must be included in the manuscript so the reviewer can assess whether the formulation was genuinely non-obvious. This is the disclosure adequacy requirement in Section~3.6. Third, even if a prior formulation was itself pipeline-generated, the framework still correctly certifies the knowledge's epistemic properties. It was designed to certify properties, not to resolve attribution, which it acknowledges as irreducible.

Routing clarification: Case~2 enters the standard track, not the
dedicated slots, because a human contribution is being claimed. Dedicated
slots are for submissions claiming no human intellectual contribution.
If a prior public formulation is cited as human contribution, the
submission enters the standard track for Layer~2 assessment. This
clarification prevents Case~2 from bypassing contribution grading.

\textit{Validation conclusion:} Layer~1 functions correctly in both
cases. Layer~2 assesses the prior formulation case without relying on
the prior disclosure claim. Attribution remains irreducible as designed.

\section{Discussion}

\subsection{Reviewer Qualification}

The framework creates an explicit requirement that did not previously
exist in stated form: reviewers must assess the pipeline-generatability
of a submission's primary contribution. The predictable editorial
challenge is this: most domain reviewers are not AI experts. The answer
is calibrated to the contribution grade being assessed, and this
calibration makes the requirement practical.

For Category~A assessment, the dedicated slot benchmark reduces the
expertise requirement to near zero. A reviewer who reads the benchmark
output for their subfield and compares a submission to it is performing
a concrete domain comparison that requires reading comprehension and
domain knowledge, not AI literacy. The benchmark does the work. This
covers the vast majority of submissions, since Category~A is the modal
case.

For Category~B assessment, the benchmark substantially helps. The
reviewer assesses whether the submission exceeds the dedicated slot
output in specific identifiable ways suggesting non-default direction at
specific stages. Domain expertise is the primary tool; the benchmark
provides the reference. The empirical finding that human-AI combinations
produce the most value when tasks require judgment that automated systems
cannot reliably provide \citep{Vaccaro2024} supports the treatment of
Category~B as a genuine contribution tier.

For Category~C assessment, the dedicated slots are necessary but not
sufficient, the slots represent what a specific pipeline configuration
produced on a specific problem set; a Category~C submission may exceed
those specific outputs because the problem set did not include this
direction. The reviewer must assess whether the pipeline would reach
this formulation from any reasonable starting point.

For Category~C papers, at least one reviewer should be explicitly
selected for pipeline familiarity in the relevant domain. This population
does not need to be imported from AI systems research. It already exists
in most pipeline-active fields: researchers who regularly use AI tools
in their own work know from direct experience what current pipeline
configurations produce.

In fields where pipelines are rarely used today, such as historical scholarship and interpretive social science, the pool of reviewers qualified for Category~C is initially small. The appropriate response is to hold off on Category~C in those fields and start with fields where pipelines are widely used. This is a rollout problem, not a flaw in the design. The current system already asks reviewers to make pipeline-generatability judgments, but it hides that requirement and leaves no way to check it. Under the proposed framework, a reviewer who rates routine Category~A work as high-impact would be misjudging pipeline-generatability; the current system makes this judgment hard because there is no tool to check oneself against and no benchmark to compare against. The proposed framework brings that judgment into the open, provides a benchmark as a reference point, and makes mistakes easy to spot, challenge, and fix. The real question is whether the framework leads to more reliable decisions using the same reviewers. For Categories~A and B, it does.

\subsection{Competitive Displacement, Equity, and Stakeholder
Implications}

The framework's implications for institutional employers, national
laboratories, large research universities, and government research agencies
differ from its implications for individual researchers. Much
institutional research is inherently Category~A and B: large-scale
computational studies, parameter sweeps, high-throughput screening,
standardized data collection. These are contributions the knowledge
enterprise genuinely needs. Category~A and B certifications are not
demotions from an unstated higher standard; they are descriptions of
what the work is in pipeline-relative terms, work the pipeline has
made more reliable and more scalable. The framework's risk for institutional employers is not that
it devalues their work but that institutional evaluation systems, if
they informally adopt A/B/C distinctions over time, might concentrate
prestige recognition on Category~C while continuing to fund and require
Category~A and B work without recognizing it as such. Addressing this
requires deliberate institutional governance decisions about how the
certification currency maps onto promotion, salary, and
mission-fulfillment criteria, decisions that lie with institutional
employers, not with the certification framework itself.

The framework serves researchers, reviewers, consumers, and institutions
differently, but the most consequential effects are on researchers
directly competing with pipeline capability. Three points on this
disruption. First: no existing researcher's publication record is
retroactively reclassified. Second: the competitive disruption is
created by the pipeline, not by the framework, naming it directly is
more useful than leaving it unnamed. Third: Category~B creates recognition
the current system cannot provide. Researchers doing genuine
human-directed pipeline work currently cannot distinguish their
contributions from automated generation in the publication record;
Category~B makes that distinction visible and certifiable, creating
recognition the current system denies them. A fourth observation applies
to the transition period: researchers submitting work now are navigating
a gap where hiring committees and fellowship panels may apply this
framework's lens informally, before any venue has officially adopted it.
The norms are shifting before the system has changed.

The equal access question has two parts. The first is affordability: researchers at under-resourced institutions may not be able to pay for access to the best pipelines. This is a real problem, but it is shrinking as API prices fall. The second is navigability: a researcher with deep domain knowledge but limited AI skills may not be able to effectively direct a pipeline, even with free access. Navigability is a skill problem that does not go away when cost does. If AI skills are unevenly spread, the framework may end up giving Category~B and C credit mostly to researchers who already know how to use AI, while those with strong domain knowledge but weak AI skills are left out. The framework itself cannot fix this. It calls for support that institutions must provide: training programs, backing for different research traditions, and funding for slow, longitudinal, rigorous work.

The equity dimensions extend beyond individual researchers globally.
Pipeline capability is concentrated in English: frontier models perform
substantially better on English-language inputs \citep{Joshi2020},
creating a capability gap the affordability-navigability framing does
not fully capture. The calibration database faces a seeding risk: if
built primarily from well-resourced institutions in wealthy countries,
in English, the contemporaneous standard will have limited reference
value in under-served domains. A researcher working on neglected
tropical diseases, informal economy dynamics in the Global South, or
indigenous ecological knowledge may find no calibration record in their
domain, not because the pipeline ceiling is low there, but because no
institution invested in building the record. In the absence of a
domain-specific record, the reviewer falls back on domain expertise
alone, as now, less calibrated, not impossible. 

A deeper question runs beneath these equity concerns: whose pipelines,
whose training data, and whose values define ``pipeline-reachable''
in the first place? The framework adjudicates contributions relative
to pipeline capability; it does not, by itself, adjudicate whose
priorities shape that capability. If frontier pipelines inherit the
linguistic and disciplinary priors of their builders, the Category~A
ceiling inherits those biases too. The framework does not dissolve
this concern. What it does is separate two questions often conflated:
whether pipeline-produced knowledge should be certified on quality
grounds (the paper's concern), and whose values should shape pipeline
capability (a political-economy question the paper deliberately does
not settle). The second is a task for science-policy institutions,
not for a certification framework.

The framework's logic applies beyond journal publication to grant applications, where credit claims directly affect who gets funded. A grant proposal faces the same question a journal reviewer does. Did the researcher's preliminary data come from a real anomaly the researcher spotted, one that pointed the pipeline toward a finding? Or did the pipeline simply run on a clearly defined problem that any skilled user could have handled? Category~B recognition gives review panels a clear way to distinguish between pipeline work a researcher directed and results generated by the pipeline on its own. The Category~A/B/C distinction fits naturally with the significance and innovation criteria that funding agencies already use.

One implication of adoption deserves a direct statement. If a substantial fraction of PhD research in pipeline-heavy empirical fields is graded Category~A, likely in systematic review, large-scale observational study, and data science, the implications for PhD training and academic labor markets are substantial. Many programs have trained researchers primarily in methodological execution: skills now widely within pipeline reach. The pipeline creates this pressure; the framework makes it legible. Institutions and funding agencies in these fields may need to consider how training models adapt to a context where methodological execution is increasingly automated, alongside their existing efforts to develop researchers' broader analytical and interpretive capacities. 

\subsection{Scope and Limits}

Three genuine limits are stated rather than resolved. Attribution uncertainty is irreducible. The framework functions well
despite it and addresses it more directly than the current system.
Overstating the framework's scope invites objections it was not designed
to address.

Adoption dynamics are unpredictable. Institutional inertia is real; the
transition sequence from adoption to norm change is a logical progression that may or
may not map onto actual institutional behavior.

A concern from researchers outside high-pipeline fields deserves direct
engagement. By making Category~A and B work visible and legible, the
framework may create conditions in which funding, doctoral recruitment,
and prestige accumulate around pipeline-tractable questions, not
because the framework rewards this, but because its infrastructure is
most easily navigated for that kind of work.

Category~C protects the specific cognitive act when it occurs. It does
not protect the broader conditions, funding for long archival
projects, support for fieldwork-intensive methods, tolerance for slow
and uncertain inquiry, that make Category~C work worth attempting in
the first place. These conditions require institutional commitments that
lie beyond what any publication framework can provide or mandate.

A scope limitation applies to research domains where frontier pipeline
capability is classified or proprietary. In national security-adjacent
fields, certain materials science applications, and other areas where
the most capable pipeline configurations are not publicly disclosed, the
dedicated slot benchmark, which is necessarily based on publicly
available systems, may systematically understate actual pipeline
capability at the frontier. Category~C certifications in these domains
carry an additional uncertainty: a contribution that exceeds the public
benchmark may nonetheless be within reach of a classified or proprietary
system the reviewer cannot consult. The contemporaneous standard cannot
resolve this from public benchmarks alone; it requires either access to
non-public capability assessments or explicit acknowledgment of the
limitation in the certification record.

A structural precondition for the framework to function as a global
standard is international coordination. Scientific publication is
already a globally integrated institution: the major venues have
international governance, international editorial boards, and authors
from every research system. A framework that adds certification
mechanisms to this institution must be designed for global coherence,
not national adoption. The contemporaneous standard depends on a shared
calibration record. Without it, the challenge mechanism becomes
unworkable across borders and the standard itself fragments into
multiple incompatible systems. Achieving
compatible standards requires coordination through existing international
scientific governance structures rather than unilateral adoption. This
coordination requirement is a necessary precondition for the framework's
full operationalization; its governance design is beyond this paper's
scope.

Three barriers stand between the framework and adoption. Reviewer capacity is resolved by calibration: the
benchmark does the work for Category~A, concentrating the AI expertise
requirement at Category~C where paper volume is lowest. First-mover
coordination is addressed by supply saturation\footnote{\textbf{Supply saturation}: the mechanism by which abundant pipeline-quality submissions raise effective acceptance thresholds at limited-slot venues, creating demand for contribution certification without requiring institutional mandate.}: venues facing increasing
volume have instrumental reasons to adopt. When one major venue admits
that pipelines can produce publishable knowledge, others face immediate
pressure to adopt or differentiate. The first commitment required is
initiation of the benchmark competition in one high-pipeline field,
a tractable step any major venue or professional society can take
independently. Institutional inertia is the hardest barrier. Universal
adoption is not required, the minimum threshold is sufficient
adoption at the evaluative frontier: prize committees, fellowship
panels, and hiring committees treating Category~C certification as a
positive signal. Adoption that begins exclusively at existing elite
frontiers will propagate existing hierarchies into the new norms;
international professional societies, with structural capacity for more
geographically representative governance, are an equally viable and
more equitable first-mover candidate.

\subsection{Partial Adoption Paths}

The framework is designed as a whole but is not all-or-nothing. The
following partial adoptions each stand on their own, each addresses a
specific failure mode in the current system, and each is within the
authority of a single decision-maker or small coalition. They are
ordered below from lowest to highest combined resistance, so that
readers in editorial, chair, or program-officer roles can act on what
is tractable without waiting for full adoption.

\textit{1. Reconsider investing in AI-detection tools.} The detection layer
is likely the wrong response to the problem.
A venue, society, or funding agency can redirect its AI-policy budget away from
detection tools and toward reviewer training without adopting any
other element of the framework. This is an immediate, self-contained
decision. The argument is in Section~1.3; the empirical backing is in
\citet{HeBu2025}, whose finding that 0.1\% of academic papers disclose
AI use despite 70\% of journals mandating it demonstrates that
detection-and-disclosure regimes have failed at scale.

\textit{2. Add pipeline-generatability to the reviewer form.} A
journal editor or conference program chair can add a single question to
the existing reviewer form: ``Could the best available automated
pipeline have generated this contribution given the existing literature at the
time of submission?'' This change can be made without adopting the Category~A/B/C
vocabulary, the dedicated slots, or anything else. The judgment is
already being made implicitly \citep{Liang2024}; making it explicit
improves review reliability immediately. One editor can make this
change. The argument is in Section~3.2 and Section~5.1.

\textit{3. Pilot dedicated benchmark slots at one conference cycle.} A
program chair at a major AI or computational-science conference can
propose piloting benchmark slots for a single cycle in one track, with
results published transparently and the governance kept separate from
the general program committee. This requires more commitment than
items~1--2 but remains within the authority of a program committee
chair acting with society or venue endorsement. The design is
specified in Section~3.4. A successful pilot provides the reference
point the full framework requires; a failed pilot is still evidence
about what pipeline-quality output looks like in that field, and the
record survives for later use.

\textit{4. Reframe significance criteria at a funding agency.} A
program officer at a national funding agency (such as the NSF, ERC, NSFC, or equivalent body) can add
pipeline-reachability language to the significance and innovation
criteria of a specific funding mechanism, without waiting for
agency-wide adoption. The change is: reviewers should assess whether
the proposed research exceeds what current automated pipelines could
generate from the problem statement and prior literature. This costs nothing
to add. It takes time to propagate through review-panel culture, but
a single mechanism or program can pilot it independently. The
grant-application argument is in Section~5.2.

\textit{5. Establish the calibration database as shared
infrastructure.} The highest-leverage but highest-resistance
adoption. A consortium of professional societies, a major publisher,
or a funding-agency coalition commits to building and maintaining the
domain-specific benchmark archive the framework relies on. This
requires multi-institution coordination and recurring funding, and
cannot be accomplished by any single actor. But every increment,
one discipline, one year, one benchmark run, creates a reference
point that did not exist before. The calibration database is the
single most important piece of infrastructure the framework lacks;
partial construction is better than none, and early seeding decisions
will shape which fields and languages are well-served a decade out.

The five paths are independent. A venue that adopts only items~1 and~2
still improves its review process materially. A conference that does
items~1, 2, and 3 has demonstrated the framework's operational
viability without requiring anyone else to act. Items~4 and~5 are
slower but disproportionately consequential when they move. 
The framework's full adoption does not depend on simultaneous action across all five.


\section{Conclusion}

The AI research pipeline separates, for the first time, two
certifications that publication has historically made simultaneously: that the
knowledge holds, and that a human made it. This paper has argued for
separating these certifications going forward: knowledge quality
through existing mechanisms adapted for pipeline-agnostic evaluation,
human contribution through pipeline-generatability grading using the
contemporaneous standard and the dedicated slot benchmark. The
decoupling is conceptually sound, implementable using
existing editorial infrastructure, and validated against the
representative hard cases.

Three real limits constrain the framework. Attribution uncertainty cannot be fully resolved. Adoption breadth is hard to predict. Category~C as a meaningful distinction depends on the persistence of cognitive acts pipelines cannot reach, a condition that may not hold indefinitely. The framework is designed to function well within these limits.

The deeper claim is not that AI has changed the nature of research, but that AI has made visible a structure that was always present. What the framework grades is the cognitive act of the researcher; the pipeline is the first public benchmark against which such acts can be identified.
More precisely: the distinction between certifying knowledge and attributing it was always available in principle but only now empirically actionable, because only now can the two come apart.
The framework proposed here grounds the recognition of human frontier contribution in the epistemic property the publication system has always been trying to certify, rather than in human origin alone.


\bibliographystyle{plainnat}
\bibliography{kc_arxiv}

\end{document}